\documentclass[letterpaper, 10 pt, conference]{ieeeconf}  

\IEEEoverridecommandlockouts                              

\usepackage{graphicx} 
\usepackage{epsfig} 
\usepackage{times} 
\usepackage{amssymb}  
\usepackage{amsmath} 
\usepackage{booktabs}
\usepackage[table]{xcolor}
\definecolor{oursrow}{RGB}{242,247,252}
\definecolor{ablaterow}{RGB}{252,247,247}
\definecolor{refrow}{RGB}{240,240,240}
\usepackage{xspace}
\usepackage{hyperref}
\usepackage{algorithm}
\usepackage{algpseudocode}

\let\labelindent\relax
\usepackage{enumitem}
\setlist{leftmargin=3.5mm}

\newcommand{\ours}{\textsc{TRACE}\xspace}
\newcommand{\oursBC}{\textsc{TRACE-BC}\xspace}
\newcommand{\teacherReplay}{\textsc{Teacher Replay}\xspace}
\newcommand{\onlineTeacher}{\textsc{Online Teacher}\xspace}

\newcommand{\bestacc}[1]{\cellcolor{blue!9}\textbf{#1}}
\newcommand{\bestsafe}[1]{\cellcolor{blue!6}\textbf{#1}}
\newcommand{\besttime}[1]{\cellcolor{green!18}\textbf{#1}}
\newcommand{\bestkey}[1]{\cellcolor{green!32}\textbf{#1}}

\title{\LARGE \bf
Plan-Conditioned Imitation for Robust Object Retrieval under Self-Occlusion in Dense Clutter 
}

\author{Kowndinya Boyalakuntla$^{1}$ \qquad
        Ajinkya Pawar$^{2}$ \qquad
        Abdeslam Boularias$^{1}$ \qquad
        Jingjin Yu$^{1}$ \\ [0.5em]
\small{$^{1}$ Rutgers University} \qquad
\small{$^{2}$ Indian Institute of Technology Bombay, India}%
}

\begin{document}

\maketitle
\thispagestyle{empty}
\pagestyle{empty}

\begin{abstract}
Retrieving objects from dense clutter requires rearrangement during which
the manipulator can occlude objects while moving them. Repeated arm
withdrawals to restore visibility interrupt execution. We introduce \ours,
a plan-conditioned imitation framework for retrieval under self-occlusion.
A single unoccluded observation initializes a digital twin, where a
privileged teacher generates a fixed nominal rollout. A recurrent student
combines local rollout context, partial object observations, and
proprioception to select actions that can correct deviations from the
prediction. Behavior cloning initializes the student; DAgger refines it
with teacher labels on student-visited states. The rollout remains fixed
throughout execution, so the deployed student needs neither online teacher
queries nor additional simulator rollouts during pushing. On 511 simulation test
scenes, \ours achieves 90.7\% success versus 43.4\% for nominal replay and
96.7\% for the privileged closed-loop teacher. At a matched 26,373-label
budget, student-state supervision achieves 87.8\% versus 66.7\% for
expert-only cloning, demonstrating gains beyond additional labels. On a
UR5e, \ours achieves 90.0\% success versus 95.0\% for the closed-loop
teacher, while reducing total execution time from 192.7\,s to 67.3\,s.
It avoids the teacher's 16.8 sensing-related arm retractions per trial
during pushing, retaining a final withdrawal for graspability evaluation. 
Code and data will be released at: \url{https://trace-retrieval.github.io/}
\end{abstract}

\section{Introduction}
\label{sec:introduction}

A target object in dense clutter may be visible yet inaccessible to a
gripper. Creating grasping clearance requires rearranging neighboring
objects, during which the robot's arm may block the camera's view while
contact continues to move objects. We address reliable target retrieval
under manipulator-induced self-occlusion while limiting interruptions for
visual-state reacquisition.
Applications include acquiring gears and electrical connectors from
assembly kit trays~\cite{nist2017assembly}, selecting books and glue bottles
for warehouse fulfillment~\cite{yu2016amazon}, and retrieving remote
controls for people with motor impairments~\cite{king2012dusty}. Such tasks
motivate creating grasping space while maintaining feedback without
repeated sensing interruptions. We study dense planar clutter with known
object footprints and a complete initial scene observation.

\begin{figure}
\centering
\includegraphics[width=1.0\linewidth]{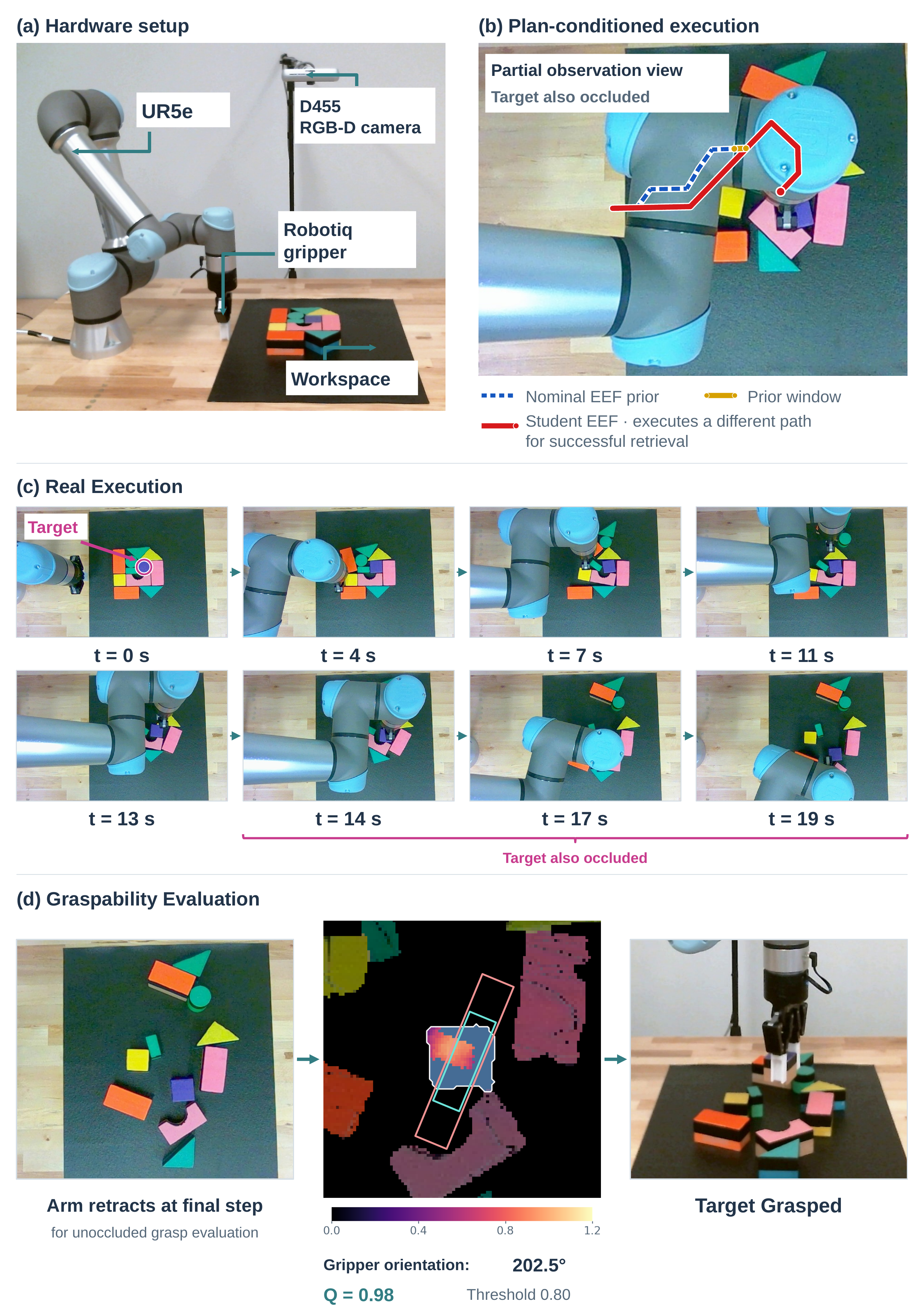}
\vspace{-6mm}
\caption{\textbf{Hardware setup and plan-conditioned execution.}
(a) UR5e with fixed external RGB-D sensing.
(b) \ours execution during target occlusion; the nominal EEF path and
active local plan window are shown for reference.
(c) Final graspability evaluation.
(d) Example pushing sequence through successful retrieval.}
\label{fig:hardware-setup}
\end{figure}
Each push changes the contacts and clearances that determine which actions
remain useful, making retrieval a long-horizon, contact-rich decision
problem. Predictive methods reason about anticipated push outcomes using
learned interaction models, tree search, or rigid-body
simulation~\cite{huang2022vft,huang2022more,huang2022pmbs}, while reactive
policies choose successive manipulations from scene
observations~\cite{kurenkov2020visuomotor,kiatos2024pregrasp}. During
self-occlusion, hidden objects may move, making earlier observations
inaccurate. Retracting the arm restores visibility but interrupts
manipulation; replaying a predicted sequence cannot correct for contact
uncertainty or execution error. The challenge is to combine prediction
with intermittent feedback to continue making useful corrective actions.

We introduce \ours, \emph{Teacher Rollouts for Adaptive Closed-loop
Execution}, a plan-conditioned imitation learning framework that combines
a fixed predictive reference, recurrent memory, and partial visual
feedback. A single unoccluded observation initializes a digital twin,
where a frozen teacher trained with complete geometric state generates a
scene-specific nominal rollout $\bar\tau$. The rollout stores predicted
end-effector positions and object centers. During execution, a GRU-based
student combines a local window from this rollout with visible object
geometry, visibility and observation-age indicators, proprioception, and
the previous action. The rollout supplies an expectation of scene
evolution, memory carries information through observation gaps, and current
detections provide evidence of execution-induced deviations. The student
predicts the complete next motion primitive and can therefore depart from
the nominal path. The rollout also supplies a scene-specific execution
budget, after which the robot checks graspability. After the initial
rollout, pushing requires neither teacher queries nor additional simulator
rollouts.

Corrective actions change the states the student subsequently encounters,
including configurations absent from teacher-controlled demonstrations.
We address this distribution shift by initializing the student with
behavior cloning and refining it using DAgger~\cite{ross2011dagger}. The
student controls execution while the frozen teacher supplies action
distributions on student-visited states. Training includes simulated
self-occlusion, detection dropout, multi-step observation blackouts, and
planning--execution mismatch. The same teacher thus provides both
pre-execution predictive context and supervision for acting in
learner-induced configurations.

The abstract summarizes the main simulation and hardware results.
Our contributions are:
\begin{itemize}
\setlength{\itemsep}{0pt}
\setlength{\parskip}{0pt}
\setlength{\parsep}{0pt}
\setlength{\topsep}{2pt}

    \item We develop \ours, which combines a one-time privileged teacher
    rollout with recurrent partial-observation control for object retrieval
    under self-occlusion.

    \item We quantify the benefit of privileged supervision on
    student-visited states using matched-label comparisons with expert-only
    behavior cloning, and evaluate recurrent memory and plan-context
    horizon.

    \item We demonstrate in simulation and on a real robot that \ours
    improves retrieval success over nominal-plan replay and reduces
    execution time relative to a complete-state teacher by avoiding
    repeated arm withdrawals for sensing during pushing.

\end{itemize}

\begin{figure}
    \centering
    \includegraphics[width=1.0\linewidth]{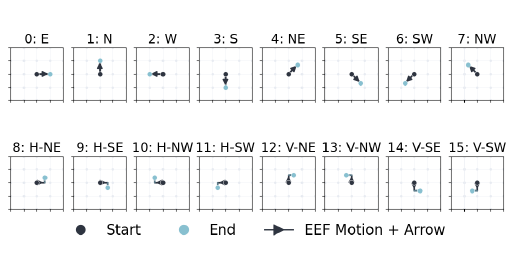}
    \vspace{-8mm}
    \caption{\textbf{End-effector motion primitives.}
The 16-action library contains four cardinal, four diagonal, and eight
two-segment staircase motions, executed horizontal-first (H-XX) or
vertical-first (V-XX).}
    \label{fig:eef_primitives}
\end{figure}
\section{Related Work}
\label{sec:related_work}

\subsection{Object Retrieval through Non-Prehensile Rearrangement}

Push-grasp methods rearrange clutter to increase grasp accessibility
\cite{zeng2018learning,xu2021efficient,berscheid2019robot}, while
Mechanical Search studies retrieval of targets hidden by clutter
\cite{danielczuk2019mechanical,kurenkov2020visuomotor}. Predictive methods reason about future accessibility:
Visual Foresight Trees searches
over outcomes predicted by a learned multi-object interaction model
\cite{huang2022vft}. MORE combines Monte Carlo tree search with
self-supervised learning to improve subsequent search
\cite{huang2022more}, while PMBS accelerates long-horizon planning through
batched rigid-body simulation \cite{huang2022pmbs}. \ours instead computes
a single pre-execution rollout as predictive context for a learned policy.

Visuomotor Mechanical Search learns closed-loop retrieval
\cite{kurenkov2020visuomotor}. Kiatos et al. also execute sequential
pre-grasp pushes without retracting for each observation
\cite{kiatos2024pregrasp}. Our distinction is rollout-conditioned recurrent
feedback with explicit missing-object observations after an initially
visible scene, rather than uninterrupted pushing alone. This also differs
from searching for a target initially hidden by clutter
\cite{xiao2019online,danielczuk2019mechanical}.
\begin{figure*}[t]
    \centering
    \vspace*{1.5mm}
    \includegraphics[width=0.9\linewidth]{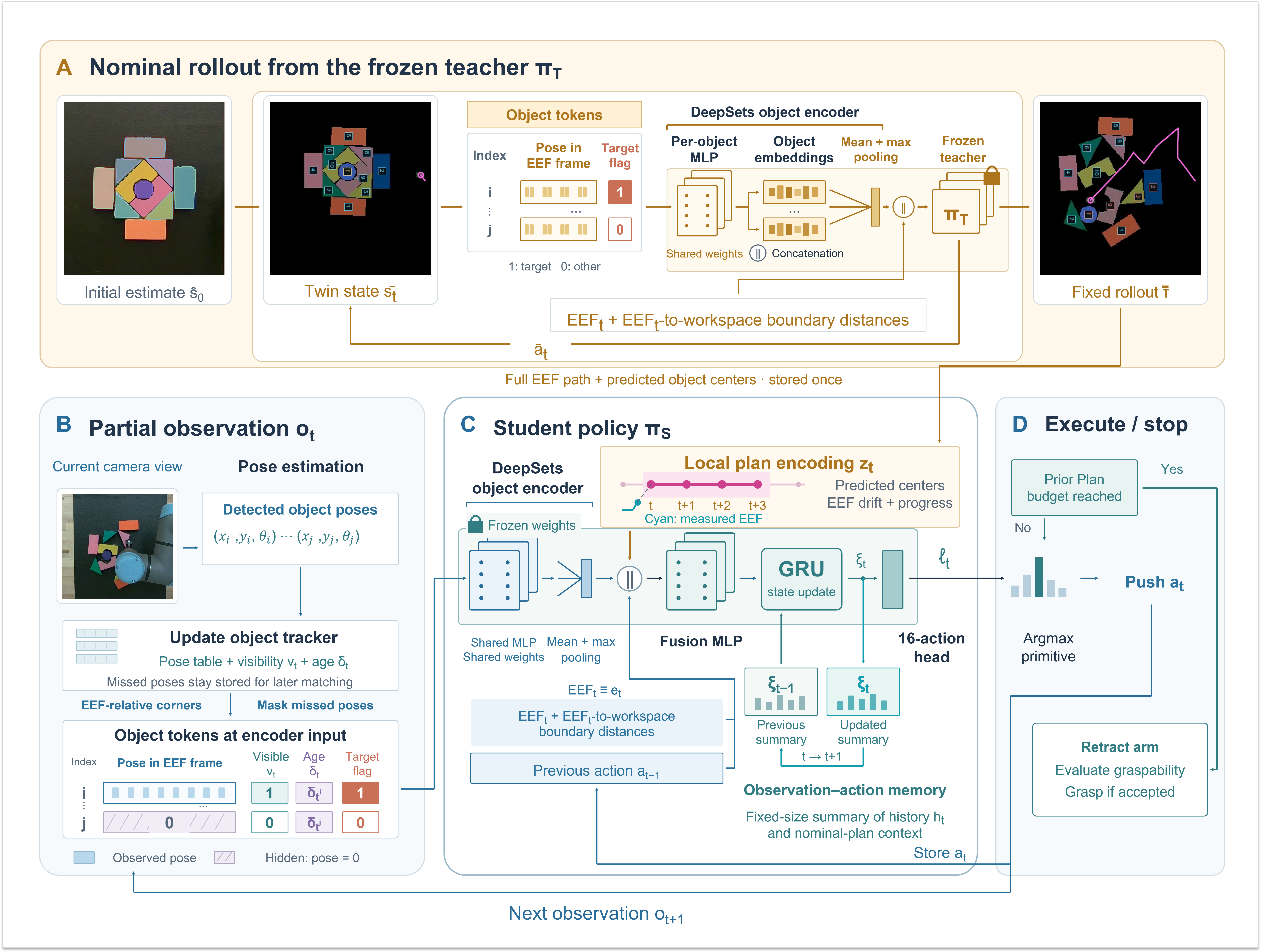}
    \caption{\textbf{\ours inference pipeline.}
    (A) A clean initial scene estimate $\hat{s}_0$ initializes a digital
    twin, where the frozen privileged teacher generates the fixed nominal
    rollout $\bar{\tau}$.
    (B) Current detections update object tracks; unavailable object
    geometry is zeroed while visibility, observation age, and target
    identity remain available.
    (C) The student combines the DeepSets scene representation, local plan
    context $z_t$, end-effector features, and previous action. A GRU
    updates recurrent state $\xi_t$, and a categorical head predicts one
    of the 16 motion primitives.
    (D) The selected primitive is executed and the next partial observation
    closes the loop. After panel (A), online execution requires neither
    teacher queries nor additional simulator rollouts.}
    \label{fig:architecture}
\end{figure*}

\subsection{Privileged Learning under Partial Observations}

Learning by Cheating distills a privileged state-based teacher into a
vision-based student \cite{chen2020cheating}. In manipulation,
Mosbach and Behnke use memory-augmented student-teacher learning for
retrieval from imperfect visual detections
\cite{mosbach2025prompt}, while VIRAL uses behavior cloning and online
DAgger to transfer privileged full-state humanoid policies to RGB-based
control \cite{he2026viral}.

Distinct teacher states can yield the same student observation but require
different actions, creating the realizability issue studied by Kim et al.
\cite{kim2025realizable}. \ours combines recurrent observation history
with a nominal rollout from the initial scene. Privileged state supplies
training labels and pre-execution context; deployment uses partial
observations and the fixed rollout.

\subsection{Imitation on Learner States and Nominal Guidance}

Behavior cloning trains on demonstrator states; DAgger addresses the
resulting covariate shift by labeling learner-induced states
\cite{ross2011dagger}. In rearrangement, one different push changes
subsequent contacts and observations. We match label budgets to isolate
the benefit of learner-state coverage using a frozen teacher.
Residual Reinforcement Learning adds a learned corrective action to the
output of a conventional controller \cite{johannink2019residual}. \ours
uses a local rollout window as input and predicts the complete next
primitive, without constraining it to a nominal action plus a residual.

\section{Problem Formulation}
\label{sec:problem}

We study target retrieval from dense planar clutter under
manipulator-induced self-occlusion. A robot rearranges neighboring
objects through non-prehensile pushes until a designated target admits a
collision-free top-down grasp. The bounded workspace
$\mathcal W\subset\mathbb R^2$ contains $n$ rigid, movable objects,
including the target. A fixed external camera
(Fig.~\ref{fig:hardware-setup}(a)) initially observes the complete scene,
from which the robot obtains an initial estimate $\hat{s}_0$. During
manipulation, however, the arm and gripper may occlude objects while
contact continues to change their poses.

We model online execution as a finite-horizon discounted partially
observable Markov decision process
$\mathcal M=(\mathcal S,\mathcal A,\mathcal T,\mathcal R,
\mathcal O,\mathcal Z,\gamma,H)$, where $\mathcal S$ and $\mathcal A$
denote the state and action spaces, $\mathcal T$ the transition model,
$\mathcal R$ the reward, $\mathcal O$ and $\mathcal Z$ the observation
space and observation model, $\gamma\in[0,1]$ the discount factor, and
$H$ the decision horizon. A privileged teacher $\pi_T$ is trained from
complete state information, whereas the deployable policy $\pi_S$ is
trained from teacher supervision and acts under partial observations.

\noindent\textbf{State Space.}
At decision step $t$, the latent state is
$s_t=(q_t^R,e_t,q_t^1,\ldots,q_t^n)$, where $q_t^R$ denotes the robot
configuration, $e_t\in\mathbb R^2$ is the planar end-effector position,
and $q_t^i\in SE(2)$, parameterized by
$(x_t^i,y_t^i,\theta_t^i)$, is the pose of object $i$. The robot
configuration determines the current collision-body poses used by the
observation model. Object footprint geometry is known and fixed.

\noindent\textbf{Action Space.}
The action space $\mathcal A$ contains the 16 discrete planar
end-effector motion primitives shown in
Fig.~\ref{fig:eef_primitives}: four cardinal, four diagonal, and eight
two-segment staircase motions. Each primitive begins at the current
end-effector position and uses a fixed commanded motion scale $\zeta$.

\noindent\textbf{Transition Function.}
Executing $a_t\in\mathcal A$ produces
$s_{t+1}\sim\mathcal T(\cdot\mid s_t,a_t)$. The transition captures
free-space end-effector motion, direct gripper--object contact, and the
resulting object--object interactions.

\noindent\textbf{Observation.}
At decision step $t$, the robot receives
$o_t=(e_t,\{\hat q_t^j\}_{j\in\mathcal V_t})\in\mathcal O$, where
$\mathcal V_t$ indexes objects detected in the current camera view and
$e_t$ remains available through proprioception. The observation model
$\mathcal Z(o_t\mid s_t)$ describes partial observations induced by
manipulator self-occlusion.

The current observation updates an object tracker initialized from
$\hat{s}_0$. For each object $i$, it maintains visibility
$v_t^i\in\{0,1\}$ and observation age $\delta_t^i$, which is reset to
zero when the object is observed and incremented otherwise. The geometry
supplied to the policy is
\[
\tilde q_t^i =
\begin{cases}
\operatorname{Rel}(\hat q_t^i,e_t), & v_t^i=1,\\
\mathbf 0, & v_t^i=0,
\end{cases}
\]
where $\operatorname{Rel}(\cdot)$ expresses planar geometry relative to
the current end effector. With target indicator $\chi^i$, the structured
observation is
\[
\tilde o_t=
\left(
e_t,
\{(\tilde f_t^i,v_t^i,\delta_t^i,\chi^i)\}_{i=1}^{n}
\right).
\]

\noindent\textbf{Reward.}
The reward $\mathcal R$ is used only to train the privileged teacher,
encouraging progress toward target graspability while maintaining
workspace containment. The student $\pi_S$ does not optimize this
reward directly; it learns from teacher supervision as described in
Sec.~\ref{sec:imitation}. The teacher reward is defined in
Sec.~\ref{sec:teacher}.

\section{Method}
\label{sec:method}
Given the initial scene estimate $\hat{s}_0$, we first roll out a frozen
privileged teacher $\pi_T$ in a digital twin to obtain a scene-specific
nominal trajectory $\bar{\tau}$. A GRU-based student then executes the
task from partial observations while conditioning on a local context
extracted from this fixed rollout. The nominal trajectory provides
predictive context rather than commands to replay, allowing the student
to depart from the predicted execution as the physical scene evolves.
Figure~\ref{fig:architecture} summarizes the inference pipeline.

\subsection{Learning a Privileged Retrieval Teacher}
\label{sec:teacher}
The privileged teacher observes the complete task geometry, all object
poses and the current end-effector state from $s_t$ and selects
actions from the same 16 motion primitives used during deployment
(Fig.~\ref{fig:eef_primitives}). Each object is represented by its
end-effector-relative planar geometry together with target indicator
$\chi^i$.

Both teacher and student use a DeepSets-style permutation-invariant
scene encoder~\cite{zaheer2017deepsets}. For object tokens
$U=\{u^i\}_{i=1}^{n}$, we define
$\mathcal E_{\theta}(U)=
[\frac{1}{n}\sum_i\psi_{\theta}(u^i),
\max_i\psi_{\theta}(u^i)]$,
where $\psi_{\theta}$ is a shared object MLP and the maximum is
componentwise. This pooling ensures permutation invariance; we make no
universality claim for this encoder.

Let $u_t^{T,i}$ denote the privileged token of object $i$. The teacher
scene representation is
$\phi_t^T=\mathcal E_{\theta_T}(\{u_t^{T,i}\}_{i=1}^{n})$.
Let $c_t$ contain the end-effector position and its distances to the
four workspace boundaries. The teacher updates
$h_t^T=\operatorname{GRU}_T(G_T([\phi_t^T,c_t]),h_{t-1}^T)$ and
produces
$p_t^T=\operatorname{softmax}(f_T(h_t^T))
=\pi_T(\cdot\mid s_t,h_{t-1}^T)$.
A separate value head supports PPO training~\cite{schulman2017ppo}.

The teacher is optimized using a graspability-based reward. Let
$g(s)\in[0,1]$ denote graspability confidence predicted by the Grasp
Network adopted from PMBS~\cite{huang2022pmbs}, and define
$\Phi(s)=2g(s)$. The reward is $r_t=10$ when
$g(s_{t+1})>0.9$ and otherwise
\begin{equation}
r_t
=
\eta_t
\left[
\gamma\Phi(s_{t+1})-\Phi(s_t)
\right]
-0.1-f_t ,
\label{eq:reward}
\end{equation}
where $\gamma=0.99$, $\eta_t=0$ on the first transition after reset and
$1$ thereafter, and $f_t$ penalizes workspace violations. The potential
difference follows reward shaping~\cite{ng1999shaping}; the initial-step
gate and terminal reward mean policy invariance is not assumed here.
The step cost favors shorter solutions. After training, the teacher is
frozen and discrete actions are selected as
$a_t^T=\arg\max_{a\in\mathcal A}p_t^T(a)$.

\subsection{Nominal Rollout and Local Plan Context}
\label{sec:nominal_context}

Before execution, $\hat{s}_0$ initializes the digital twin and the frozen
teacher is rolled out once (Fig.~\ref{fig:architecture}(A)). We retain
the nominal trajectory
$\bar{\tau}=\{(\bar e_j,\bar P_j)\}_{j=0}^{N-1}$, where
$\bar P_j=\{\bar p_j^i\}_{i=1}^{n}$ contains the predicted object
centers. The rollout remains fixed throughout the corresponding
execution episode.
At student decision $t$, let $j_k=\min(t+k,N-1)$ for
$k\in\{0,1,2,3\}$. The local plan context is
$z_t=[\{\bar e_{j_k}-e_t\}_{k=0}^{3},
\rho_t,\beta_t,\{\bar p_{j_0}^i-e_t\}_{i=1}^{n}]$, where
$\rho_t=\min(1,t/\max(1,N-1))$ denotes rollout progress and
$\beta_t=\mathbf{1}[t\geq N]$ indicates nominal-rollout exhaustion. Requested indices beyond the stored rollout reuse its
final sample.

All nominal positions are expressed relative to the current measured
end-effector position. Thus, $\bar{\tau}$ remains fixed while $z_t$
changes with physical execution. The student receives geometric plan
context rather than the teacher's nominal primitive identities.

\subsection{Plan-Conditioned Recurrent Student}
\label{sec:student}

At each decision, the structured observation $\tilde o_t$ defined in
Sec.~\ref{sec:problem} provides one token for every tracked object
(Fig.~\ref{fig:architecture}(B)). Visible objects contribute their
current end-effector-relative geometry; unavailable geometry is zeroed,
while visibility, observation age, and target identity remain encoded.
For object token $u_t^{S,i}$, the student scene representation is
$\phi_t^S=\mathcal E_{\theta_S}(\{u_t^{S,i}\}_{i=1}^{n})$, using the
same mean--max aggregation as the teacher but separate parameters.
The instantaneous network input is
$x_t=[\phi_t^S,c_t,z_t,\operatorname{onehot}(a_{t-1})]$.
A fusion MLP produces $u_t=G_S(x_t)$, and the recurrent state updates as
$\xi_t=\operatorname{GRU}_S(u_t,\xi_{t-1})$
(Fig.~\ref{fig:architecture}(C)).
The recurrent state allows information from earlier observations and
actions to influence decisions when current object geometry is
unavailable. It is not trained to explicitly reconstruct the hidden
scene; instead, it provides task-relevant temporal context for action
selection. The categorical action head produces
$p_t^S=\operatorname{softmax}(f_S(\xi_t))
=\pi_S(\cdot\mid x_t,\xi_{t-1})$, and the executed primitive is
$a_t^S=\arg\max_{a\in\mathcal A}p_t^S(a)$.

\begin{algorithm}[t]
\caption{\ours Student Training}
\label{alg:training}
\scriptsize
\algrenewcommand\algorithmicindent{1em}
\begin{algorithmic}[1]
\Require Training scenes $\mathcal S_{\mathrm{train}}$,
frozen teacher $\pi_T$, DAgger rounds $K$

\State $\mathcal D_0 \gets \varnothing$

\For{each scene in $\mathcal S_{\mathrm{train}}$}
    \State $\bar\tau \gets \Call{NominalRollout}{\pi_T}$
    \For{each teacher-visited state $s_t$}
        \State $\tilde o_t \gets \Call{PartialObs}{s_t}$
        \State $x_t \gets
        [\phi_t^S,c_t,z_t,\operatorname{onehot}(a_{t-1})]$
        \State $\mathcal D_0 \gets
        \mathcal D_0 \cup \{(x_t,p_t^T)\}$
    \EndFor
\EndFor

\State $\pi_S^0 \gets \Call{FitStudent}{\mathcal D_0}$

\For{$k=1,\ldots,K$}
    \State $\mathcal B_k \gets \varnothing$
    \For{each scene in $\mathcal S_{\mathrm{train}}$}
        \State $\bar\tau \gets \Call{NominalRollout}{\pi_T}$
        \For{each state $s_t$ visited by $\pi_S^{k-1}$}
            \State $\tilde o_t \gets \Call{PartialObs}{s_t}$
            \State $x_t \gets
            [\phi_t^S,c_t,z_t,\operatorname{onehot}(a_{t-1})]$
            \State $p_t^T \gets
            \pi_T(\cdot\mid s_t,h_{t-1}^T)$
            \State $\mathcal B_k \gets
            \mathcal B_k \cup \{(x_t,p_t^T)\}$
        \EndFor
    \EndFor
    \State $\mathcal D_k \gets \mathcal D_{k-1}\cup\mathcal B_k$
    \State $\pi_S^k \gets \Call{FitStudent}{\mathcal D_k}$
    \Comment{Initializes a new student and trains it on $\mathcal D_k$
    using Eq.~\eqref{eq:student_loss}}
\EndFor

\State \Return student selected on held-out validation scenes

\vspace{0.4em}
\Function{PartialObs}{$s_t$}
    \State $\bar v_t^i \gets
    \mathbf{1}[F_t^i \cap M_t^R=\varnothing]$
    \State $d_t^i\sim\operatorname{Bernoulli}(p_{\mathrm{drop}}),
    \quad
    \kappa_t=\mathbf{1}[t\in\mathcal B_{\mathrm{out}}]$
    \State $v_t^i\gets
    (1-\kappa_t)\,\bar v_t^i(1-d_t^i),\quad \forall i$
    \State $\delta_t^i\gets
    (1-v_t^i)(\delta_{t-1}^i+1),
    \quad
    \tilde q_t^i\gets
    v_t^i\operatorname{Rel}(\hat q_t^i,e_t)$
    \State \Return
    $\tilde o_t=
    (e_t,\{(\tilde q_t^i,v_t^i,\delta_t^i,\chi^i)\}_{i=1}^{n})$
\EndFunction
\end{algorithmic}
\end{algorithm}

\subsection{Learning from Privileged Supervision}
\label{sec:imitation}

We train the student in two stages using supervision from the frozen
teacher. For each training scene, the teacher first generates the fixed
nominal rollout $\bar{\tau}$ from the nominal scene. Student trajectories
are then collected in the corresponding execution scene, which may be
perturbed while $\bar{\tau}$ remains unchanged.
During data collection, we simulate manipulator-induced self-occlusion
using the current robot geometry. For collision body $b$, let $V^b$
denote its collision-mesh vertices and $T_t^b$ its current pose. Its
planar occlusion region is
$M_t^b=\operatorname{Rect}_{\epsilon}
(\operatorname{Hull}(\Pi_{xy}(T_t^bV^b)))$, and
$M_t^R=\bigcup_b M_t^b$.
Here, $\Pi_{xy}$ projects transformed collision geometry onto the
workspace plane and $\operatorname{Rect}_{\epsilon}$ is a minimum-area
enclosing rectangle padded by $\epsilon=10$\,mm. With object footprint
$F_t^i$, self-occlusion visibility is
$\bar v_t^i=\mathbf{1}[F_t^i\cap M_t^R=\varnothing]$.
This provides a conservative planar approximation of manipulator
self-occlusion rather than an exact rendering of the camera silhouette.
Training additionally applies independent token dropout
$d_t^i\sim\operatorname{Bernoulli}(p_{\mathrm{drop}})$ and scheduled
multi-decision blackouts with indicator
$\kappa_t=\mathbf{1}[t\in\mathcal B_{\mathrm{out}}]$.
The resulting visibility is
$v_t^i=(1-\kappa_t)\bar v_t^i(1-d_t^i)$.
Unavailable object geometry is zeroed, while observation age
$\delta_t^i$ is reset when $v_t^i=1$ and incremented otherwise.
Multi-step blackouts expose the GRU to consecutive decisions without
current object geometry, encouraging the policy to exploit temporal
context together with proprioception and nominal-plan context. We denote
the resulting structured observation by $\tilde o_t=\Omega(s_t)$. Algorithm~\ref{alg:training} summarizes the complete student-training
procedure.

\noindent\textbf{Behavior cloning.}
The initial dataset $\mathcal D_0$ is collected from teacher-controlled
trajectories. At every visited state, the teacher acts from privileged
state information, while the student input is constructed from
$\tilde o_t$, the fixed-rollout context $z_t$, end-effector features
$c_t$, and the previous action. Each recurrent sequence therefore pairs
the information available to the student with the teacher's categorical
action distribution $p_t^T$. Training on $\mathcal D_0$ yields the
behavior-cloned policy $\pi_S^0$.

\noindent\textbf{DAgger.}
Behavior cloning exposes the student only to states visited under teacher
control. At DAgger round $k$, the current student $\pi_S^{k-1}$ instead
controls execution, while the frozen teacher labels every student-visited
state with $p_t^T$. The student's selected primitive remains the executed
action; the teacher output is used only as the supervised target. The
teacher recurrent state $h_t^T$ is advanced along the same student-induced
state sequence.
Let $\mathcal B_k$ denote the recurrent sequences collected at round
$k$. We aggregate
$\mathcal D_k=\mathcal D_{k-1}\cup\mathcal B_k$
and fit the next student on $\mathcal D_k$. Thus, DAgger preserves the
same privileged supervision used for behavior cloning while extending it
to states induced by the student's own actions.
Across behavior cloning and DAgger, the student minimizes a
recovery-weighted forward-KL objective,
\begin{equation}
\mathcal L_{\mathrm{IL}}
=
\frac{
\sum_t m_t w_t
D_{\mathrm{KL}}\!\left(p_t^T\|p_t^S\right)
}{
\sum_t m_t w_t
},
\label{eq:student_loss}
\end{equation}
where the sums extend over the sampled recurrent sequences.
Here, $p_t^T$ and $p_t^S$ are the teacher and student categorical
distributions over the 16 motion primitives. The mask $m_t$ selects
valid action-supervision steps, excluding padding, terminal observations,
invalid states, and states already satisfying the graspability criterion;
occluded observations remain eligible for supervision. The recovery weight $w_t\in[1,3]$ increases the contribution of decisions
for which recovery margin is limited, based on progress toward the nominal
horizon, remaining step slack, and remaining travel slack. These quantities
depend only on the recorded execution and nominal rollout, not on future
success labels. Normalization by $\sum_t m_t w_t$ prevents the weighting
magnitude from rescaling the overall loss. Matching the full teacher
distribution preserves relative preferences among corrective actions
beyond the argmax.

\section{Experiments}
\label{sec:experiments}

We evaluate four questions: whether plan-conditioned closed-loop
execution improves over nominal replay and existing retrieval methods;
whether supervision on student-visited states provides benefit beyond
additional expert data; how recurrent memory and the plan-context horizon
affect performance under partial observation; and whether these advantages
transfer to physical retrieval without repeated complete-scene
reacquisition.

\subsection{Experimental Protocol}
\label{sec:exp_protocol}

\begin{figure}[t]
    \centering
    \includegraphics[width=1.0\linewidth]{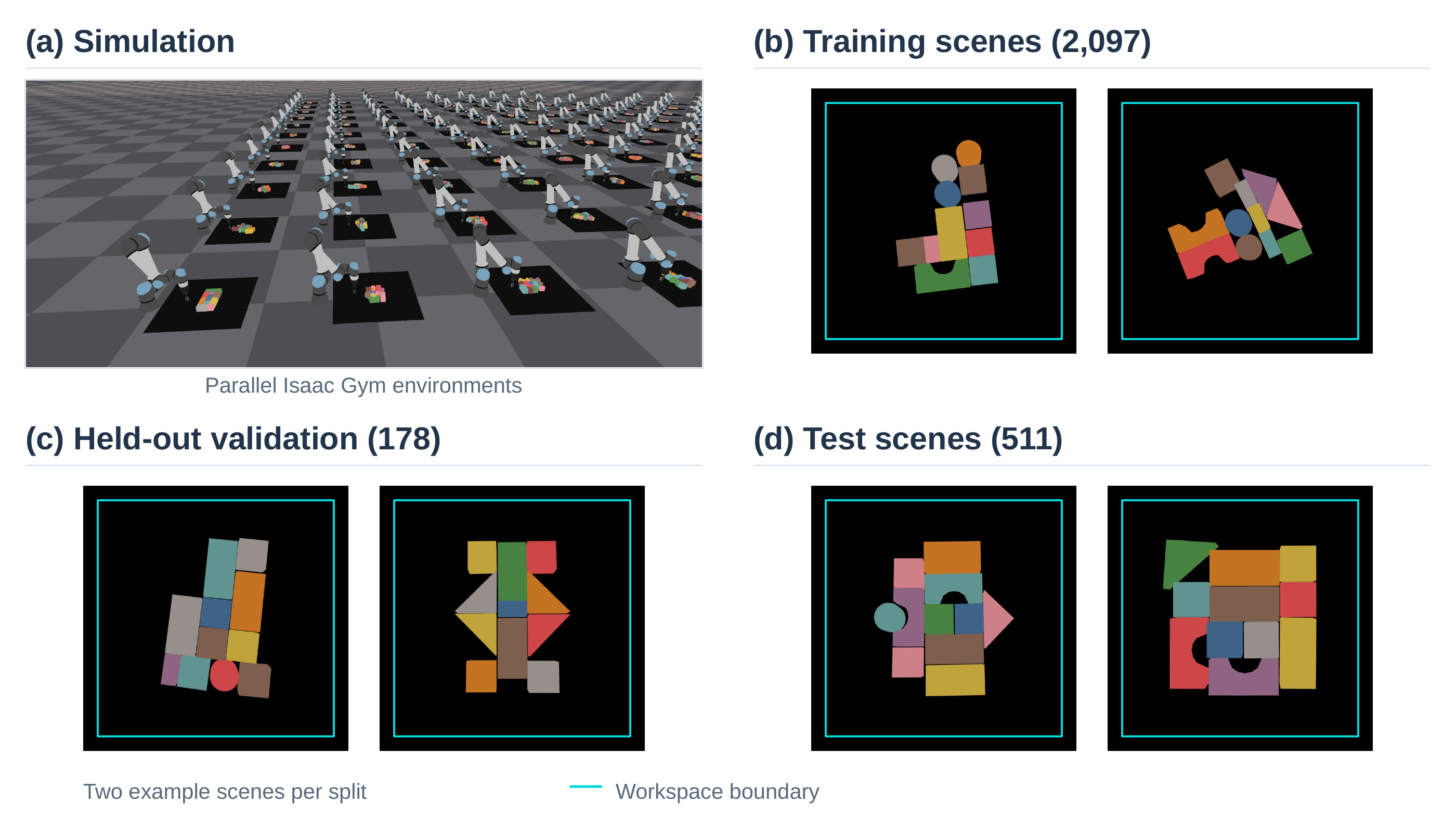}
   \caption{\textbf{Simulation setup and dataset.}
(a) Parallel Isaac Gym environments. (b)-(d) Representative training,
validation, and test scenes (2,097/178/511); validation is used for model
selection and the disjoint test set only for final evaluation.}
    \label{fig:simulation_dataset}
\end{figure}

\noindent\textbf{Environment and scenes.}
Experiments use Isaac Gym~\cite{makoviychuk2021isaac} with eleven movable objects in a
$0.448\times0.448$\,m workspace and the 16 primitives in
Fig.~\ref{fig:eef_primitives}. We use 2,097 training, 178 held-out
validation, and 511 independently generated test scenes; validation is
used for model selection and the test set only for final evaluation
(Fig.~\ref{fig:simulation_dataset}).

\noindent\textbf{Planning--execution mismatch.}
For each episode, $\bar\tau$ is generated from the initial planning
scene, after which every object in the corresponding execution scene is
perturbed independently by up to $\pm15$\,mm in each planar direction
and $\pm10^\circ$ in yaw. Thus, the nominal rollout predicts a nearby
scene rather than revealing the executed state. Paired policies receive
identical perturbations.

\noindent\textbf{Partial-observation condition.}
Manipulator self-occlusion follows the geometric model in
Sec.~\ref{sec:imitation}. During training, otherwise visible object
tokens are additionally dropped with probability
$p_{\mathrm{drop}}=10\%$, and scheduled blackouts remove all object
geometry for 5 consecutive decisions. 
Visible objects use exact simulator geometry so that these experiments
isolate missing observations from pose-estimation error. At test time, the same self-occlusion model is active, together with
$p_{\mathrm{drop}}=10\%$ token dropout at every decision and one
5-decision blackout whose onset is sampled uniformly over the
120-decision horizon. Paired policies receive identical corruption draws.

\noindent\textbf{Training.}
The teacher is trained with PPO using Eq.~\eqref{eq:reward} and then
frozen. The student is initialized by behavior cloning and refined with
DAgger (Sec.~\ref{sec:imitation}). Each student fit uses 10,000 optimizer
updates with minibatches of 32 recurrent sequences and minimizes
Eq.~\eqref{eq:student_loss}. Checkpoints and the DAgger round are selected using the 178 held-out validation scenes. The 511 test scenes are not used for model selection.

\noindent\textbf{Metrics and statistics.}
An episode succeeds when predicted graspability exceeds $0.9$ while all
objects remain inside the workspace. We separately report out-of-workspace (OOW) and \emph{Budget} failures,
the latter indicating that a method reached its execution limit before
success. \ours and \oursBC use the scene-specific teacher-relative
step/travel budget, without additional travel tolerance in simulation;
\teacherReplay executes the recorded nominal sequence to completion.
The remaining methods use their predefined timeout or action limits. Success
confidence intervals use 2,000 stratified scene-bootstrap resamples;
paired comparisons use identical scenes and initial perturbations.

\noindent\textbf{Reference policies and baselines.}
\teacherReplay executes the teacher-generated nominal sequence without
online object feedback. \oursBC uses the same recurrent
plan-conditioned architecture as \ours but is trained only by behavior
cloning. \onlineTeacher executes the privileged teacher from complete
state. We additionally compare against PMBS and Serial MCTS
\cite{huang2022pmbs}, which require complete scene state, and the
target-centric Spiral and Straight-Line heuristics. The information
available to each method is summarized in Table~\ref{tab:sim_main}.

\begin{figure}
    \centering
    \includegraphics[width=1.0\linewidth]{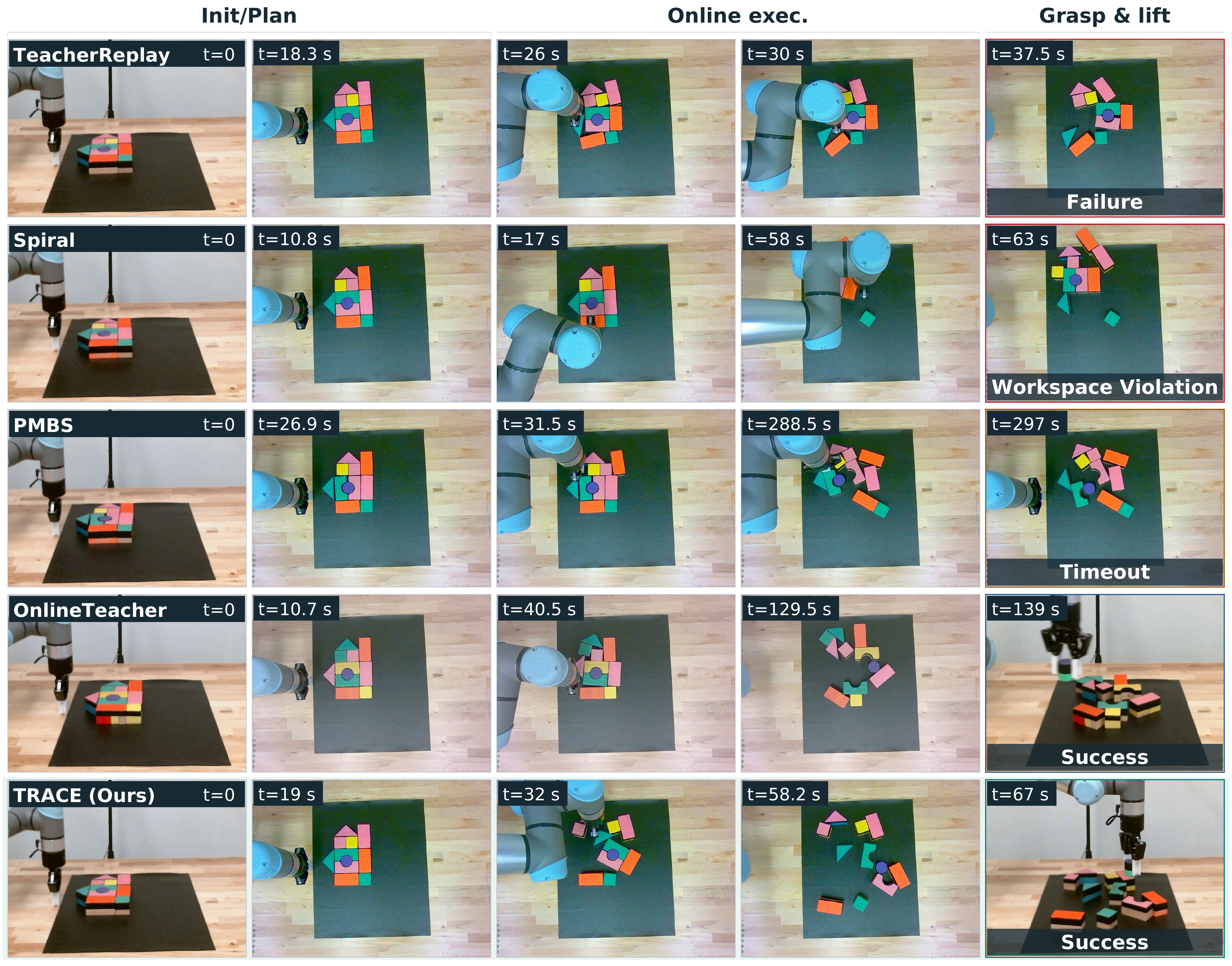}
    \caption{\textbf{Qualitative real-robot comparison.}
Representative outcomes: \teacherReplay ends ungraspable, Spiral causes
a workspace exit, PMBS reaches its 15-push limit, \onlineTeacher succeeds
after repeated complete-scene reacquisition, and \ours succeeds without
online arm retraction. Timestamps are in seconds.}
    \label{fig:hardware_qualitative}
\end{figure}
\subsection{Plan-Conditioned Execution and Reference Policies}
\label{sec:sim_main}

We compare plan-conditioned closed-loop execution against open-loop
replay of the same nominal rollout and other retrieval references.
Table~\ref{tab:sim_main} compares \ours with its internal references,
planning-based baselines, target-centric heuristics, and the privileged
\onlineTeacher.
\begin{table}[t]
\centering
\caption{\textbf{Simulation comparison} on the 511-scene test set.
OOW denotes a workspace violation; Budget denotes the method-specific
execution limit. Brackets are 95\% stratified scene-bootstrap CIs
from 2,000 resamples.}
\label{tab:sim_main}

\scriptsize
\setlength{\tabcolsep}{2.5pt}
\renewcommand{\arraystretch}{1.06}

\resizebox{\columnwidth}{!}{%
\begin{tabular}{@{}llcccc@{}}
\toprule
Method
& Online state
& Success
& 95\% CI
& OOW
& Budget \\
&
& (\%)
&
& (\%)
& (\%) \\
\midrule

\multicolumn{6}{@{}l}{\textit{Nominal-rollout methods}} \\

\teacherReplay
& None
& 43.4
& [39.1,48.0]
& 6.3
& 50.3 \\

\oursBC
& Partial
& 66.7
& [63.7,69.8]
& 15.3
& 18.0 \\

\rowcolor{oursrow}
\textbf{\ours}
& Partial
& \textbf{90.7}
& [88.9,92.6]
& 7.2
& 2.1 \\

\midrule
\multicolumn{6}{@{}l}{\textit{Planning and heuristic baselines}} \\

PMBS
& Complete
& 88.1
& [85.1,90.8]
& ---
& 11.9 \\

Serial MCTS
& Complete
& 44.0
& [40.1,48.1]
& ---
& 56.0 \\

Spiral
& Target position
& 80.2
& [76.9,83.8]
& 19.6
& 0.2 \\

Straight Line
& Target position
& 0.4
& [0.0,1.0]
& 0.0
& 99.6 \\

\midrule
\multicolumn{6}{@{}l}{\textit{Privileged reference}} \\

\rowcolor{refrow}
\onlineTeacher
& Complete
& \textbf{96.7}
& [95.1,98.0]
& 3.3
& 0.0 \\

\bottomrule
\end{tabular}%
}
\end{table}

\teacherReplay succeeds in only 43.4\% of scenes, whereas \ours reaches
90.7\% using the same nominal prediction together with partial
closed-loop observations. \ours approaches \onlineTeacher (96.7\%) and
also exceeds PMBS (88.1\%) despite using less online state information.
The lower \oursBC performance (66.7\%) motivates the student-state
supervision study in Sec.~\ref{sec:dagger_results}.

\subsection{Effect of Student-State Supervision}
\label{sec:dagger_results}

\begin{table}[t]
\centering
\caption{\textbf{Effect of student-state supervision} on the 511-scene
test set. Results average three seeds; all fits use 10,000 updates.
$\Delta$ is relative to the preceding round in the aggregation block
and Expert BC in the matched-budget blocks; brackets are paired 95\% CIs.}
\label{tab:dagger}

\scriptsize
\setlength{\tabcolsep}{2.8pt}
\renewcommand{\arraystretch}{1.08}

\resizebox{\columnwidth}{!}{%
\begin{tabular}{@{}lrrc@{}}
\toprule
Training data
& Labels
& Success
& $\Delta$ \\
&
&
(\%)
& (pp, 95\% CI) \\
\midrule

\multicolumn{4}{@{}l}{\textit{DAgger aggregation}} \\

Expert BC
& 26,373
& 66.7
& -- \\

DAgger R1
& 57,125
& 87.1
& $+20.4\;[17.5,23.2]$ \\

DAgger R2
& 84,429
& 89.0
& $+2.0\;[-0.1,4.0]$ \\

\rowcolor{oursrow}
\textbf{\ours\ (DAgger R3)}
& 111,555
& \textbf{90.7}
& $+1.7\;[-0.2,3.5]$ \\

\midrule

\multicolumn{4}{@{}l}{\textit{Matched label budget}} \\

\rowcolor{refrow}
Expert BC
& 26,373
& 66.7
& -- \\

\rowcolor{ablaterow}
Student-state BC
& 26,373
& \textbf{87.8}
& $\mathbf{+21.1}\;[18.3,23.9]$ \\

\midrule

\multicolumn{4}{@{}l}{\textit{Matched full-data budget}} \\

\rowcolor{refrow}
Expert BC, full data
& 111,555
& 78.8
& -- \\

\rowcolor{oursrow}
\textbf{\ours\ (DAgger R3)}
& 111,555
& \textbf{90.7}
& $\mathbf{+11.9}\;[9.5,14.4]$ \\

\bottomrule
\end{tabular}%
}
\end{table}

We test whether DAgger helps because supervision is collected on
student-visited states rather than simply because aggregation provides
more labels. Expert BC uses 2,097 teacher-controlled trajectories
containing 26,373 valid labels; three DAgger rounds then progressively
add teacher labels on states induced by the current student.
Most of the gain appears after the first aggregation round, with later
rounds giving smaller improvements. More importantly, the matched-label
control improves success from 66.7\% to 87.8\% at the same 26,373-label
and optimization budgets. Even after increasing Expert BC to the full
111,555-label budget, it reaches only 78.8\%, compared with 90.7\% for
DAgger R3. Together, these controls show that student-state coverage,
rather than additional supervision alone, accounts for most of the
DAgger advantage.

\subsection{Plan-Context Horizon and Recurrent Memory}
\label{sec:architecture_ablations}

\begin{table}[t]
\centering
\caption{\textbf{Plan-context horizon and recurrent-memory ablations}
on the 511-scene test set (three seeds). $\Delta$ and 95\% CIs are
paired against \ours ($K=4$); Budget is the teacher-relative
step/travel limit. Steps is the mean over successful episodes.}
\label{tab:architecture_ablation}

\scriptsize
\setlength{\tabcolsep}{2.0pt}
\renewcommand{\arraystretch}{1.08}

\resizebox{\columnwidth}{!}{%
\begin{tabular}{@{}lccccc@{}}
\toprule
Variant
& Success
& $\Delta$ vs. \ours
& OOW
& Budget
& Steps \\
& (\%)
& (pp, 95\% CI)
& (\%)
& (\%)
& \\
\midrule

\rowcolor{refrow}
\textbf{\ours ($K=4$)}
& \textbf{90.7}
& ---
& 7.2
& 2.1
& 14.8 \\

\rowcolor{ablaterow}
w/o GRU
& 85.8
& $\mathbf{-4.9}\;[-6.5,-2.4]$
& 8.0
& \cellcolor{orange!18}\textbf{6.2}
& 15.0 \\

\midrule

$K=1$
& 88.5
& $-2.2\;[-3.7,+0.2]$
& 8.4
& 3.1
& 14.6 \\

$K=2$
& 89.9
& $-0.8\;[-2.1,+1.4]$
& 7.8
& 2.3
& 14.7 \\

$K=8$
& 89.3
& $-1.4\;[-2.7,+0.8]$
& 7.7
& 3.0
& 14.7 \\

Full plan
& 90.0
& $-0.7\;[-2.0,+1.6]$
& 7.3
& 2.6
& 14.4 \\

\bottomrule
\end{tabular}%
}
\end{table}
Among plan-conditioned policies, we ablate recurrent memory and the
nominal-plan context horizon while holding the remaining protocol fixed.
We do not treat a plan-free controller as a matched ablation because
removing $\bar\tau$ also removes the scene-specific execution horizon and
therefore introduces a separate termination-policy design choice.
Removing the GRU primarily increases execution-budget exhaustion, while
the OOW rate remains nearly unchanged, supporting the role of recurrence
in carrying useful temporal context through missing observations.
In contrast, performance is similar across the tested plan horizons, and
the full nominal rollout provides no observed benefit over $K=4$.
We therefore retain $K=4$ as a compact local context.

\begin{table*}[t]
\centering
\caption{\textbf{Real-robot evaluation} on 20 scenes with two trials
each. OOW denotes a workspace violation. Arm retractions count
visual-state reacquisition; \ours's final graspability-check retraction
is excluded. Other unsuccessful trials reached the method-specific
execution limit.}

\label{tab:hardware_main}

\scriptsize
\setlength{\tabcolsep}{2.3pt}
\renewcommand{\arraystretch}{0.95}

\resizebox{\textwidth}{!}{%
\begin{tabular}{lccrrrrrrr}
\toprule
Method & Execution & Online information
& Success (\%) & OOW (\%)
& Init./plan (s) & Online exec. (s) & Grasp+lift (s)
& Total (s) & Arm retractions \\
\midrule

\teacherReplay & Open loop & Initial plan only
& 47.5
& \bestsafe{0.0}
& 27.5 & 11.6 & 10.4
& 49.5
& \bestkey{0.0} \\

Spiral & Closed loop & Target pose
& 77.5
& 17.5
& \besttime{6.3}
& 29.2
& 8.9
& \bestkey{44.4}
& 2.8 \\

PMBS & Closed loop & Complete scene
& 85.0
& 10.0
& 124.9
& \besttime{7.4}
& \besttime{8.8}
& 141.1
& 3.9 \\

\onlineTeacher & Closed loop & Complete scene
& \bestacc{95.0}
& \bestsafe{0.0}
& 10.4
& 173.3
& 9.0
& 192.7
& 16.8 \\

\ours (Ours) & Closed loop & Partial scene + $\bar\tau$
& 90.0
& \bestsafe{0.0}
& 30.4
& 24.8
& 12.1
& 67.3
& \bestkey{0.0} \\

\bottomrule
\end{tabular}%
}
\end{table*}

\subsection{Real-Robot Evaluation}
\label{sec:hardware_results}

We evaluate whether the simulation-trained policy transfers to physical
retrieval and whether partial-observation execution reduces the overhead
of obtaining complete scene state. The system uses a UR5e, Robotiq
parallel-jaw gripper, and fixed Intel RealSense D455
(Fig.~\ref{fig:hardware-setup}). A clean initial observation initializes
the digital twin and generates $\bar\tau$ once before manipulation.

\noindent\textbf{Protocol.}
The benchmark contains 20 unseen clutter arrangements with two trials per
scene and method. Success requires the target to become graspable, be
successfully grasped and lifted, and remain within workspace constraints.
We report success, OOW failures, initialization/planning time, online
execution time, grasp-and-lift time, total time, and deliberate arm
retractions used for visual-state reacquisition. For PMBS and \onlineTeacher, these retractions restore complete scene
state; for Spiral, they are required only when the target pose is
occluded.
The fixed camera continues to provide partial observations throughout
execution without requiring arm motion. In contrast, methods that require
the \emph{complete} scene state must retract the arm to restore visibility;
we count each such arm-retraction event as a complete-scene
reacquisition. The final arm retraction used by \ours for terminal
graspability evaluation is not counted as an online reacquisition. 
The \ours policy has no learned stop action; pushing terminates at the
scene-specific teacher-relative step/travel budget with an additional
5\,cm travel tolerance. Other methods use their predefined action or
timeout limits. Table~\ref{tab:hardware_main} reports success and OOW;
remaining unsuccessful trials terminate at the corresponding
method-specific execution limit.

\noindent\textbf{References.}
\teacherReplay uses only the initial nominal plan. Spiral requires the
target pose at each decision: while the target remains visible, this pose
is updated directly from the fixed camera without arm retraction; when
the target is occluded, the arm retracts to reacquire it. PMBS and
\onlineTeacher require complete scene state and therefore retract the arm
whenever full visibility must be restored.

\noindent\textbf{Results.}
Figure~\ref{fig:hardware_qualitative} illustrates representative
execution behaviors and failure modes across the evaluated methods.
\ours reaches 90.0\% hardware success without any arm retraction for
state reacquisition during pushing, compared with 47.5\% for
\teacherReplay. Spiral has the lowest total time at 44.4\,s and requires
only 2.8 retraction-based reacquisitions per trial on average, but reaches
77.5\% success and incurs OOW failures in 17.5\% of trials. Spiral shows that target-only feedback can be obtained with relatively
little reacquisition overhead, but this heuristic remains less reliable
than \ours and incurs substantially more OOW failures. \onlineTeacher reaches 95.0\% success, but requires 16.8 arm retractions
per trial to recover complete scene state and 192.7\,s total execution
time, versus 67.3\,s for \ours. PMBS similarly requires complete-state
reacquisition and reaches 85.0\% success with 141.1\,s total time.
Overall, \ours retains much of the reliability of privileged
closed-loop execution while avoiding the repeated arm-withdrawal overhead
needed to restore complete scene visibility.

\section{Conclusions}
We presented \ours, which uses a one-time privileged nominal rollout
as predictive context for recurrent retrieval under self-occlusion.
The design retains much of privileged closed-loop performance while
avoiding repeated complete-scene reacquisition; on hardware, the rollout
also provides a scene-specific execution horizon without a learned stop
action. Nominal plans thus support feedback under partial observability.
Matched-label comparisons show that supervision on student-visited states
improves retrieval reliability beyond adding teacher demonstrations alone. Architecture
ablations further support combining recurrent memory with local rollout
context to respond to deviations during contact. The current evidence is limited to planar
clutter with known object footprints and an initially visible scene.
Uncertain initial estimates and unfamiliar geometries require further evaluation.
An important next step is detecting when the nominal rollout becomes
uninformative and selectively acquiring a new view, balancing sensing
overhead against the risk of continued execution.
Our implementation uses discrete motion primitives and a
teacher-relative budget; future work will study continuous end-effector
actions, learned termination, and object retrieval from non-planar (3D) clutter.







\bibliographystyle{IEEEtran}
\bibliography{references}

\end{document}